# A New Neuromorphic Computing Approach for Epileptic Seizure Prediction

Fengshi Tian[1,2], Jie Yang[1], Shiqi Zhao[1], Mohamad Sawan[1], *Fellow, IEEE*

[1]CenBRAIN Lab., School of Engineering, Westlake University, Hangzhou, Zhejiang, China 310024

[2]School of Microeletronics, Fudan University, Shanghai, China 200433

Email: yangjie@westlake.edu.cn

*Abstract*—Several high specificity and sensitivity seizure prediction methods with convolutional neural networks (CNNs) are reported. However, CNNs are computationally expensive and power hungry. These inconveniences make CNN-based methods hard to be implemented on wearable devices. Motivated by the energy-efficient spiking neural networks (SNNs), a neuromorphic computing approach for seizure prediction is proposed in this work. This approach uses a designed gaussian random discrete encoder to generate spike sequences from the EEG samples and make predictions in a spiking convolutional neural network (Spiking-CNN) which combines the advantages of CNNs and SNNs. The experimental results show that the sensitivity, specificity and AUC can remain 95.1%, 99.2% and 0.912 respectively while the computation complexity is reduced by 98.58% compared to CNN, indicating that the proposed Spiking-CNN is hardware friendly and of high precision.

## I. INTRODUCTION

Epilepsy is one of the most common neurological diseases worldwide. Up to 35% of around 60 million epileptic patients are not receiving effective medical treatment because of being drug-refractory [1], [2], [3]. Those patients may suffer from severe co-morbidities, injuries and anxiety due to sudden seizure onset [4]. Therefore, it is significant to propose a highly accurate and energy-efficient approach to predict seizure onsets. Electroencephalography (EEG) is an electrophysiological technique for the recording of the electrical activity of brains. Representing brain activity of epileptic patient, EEG is commonly used for epileptic seizure prediction [5]. Typical EEG recordings of an epileptic patient can be defined as four states: Interictal (between seizures), Preictal (before seizure), Ictal (seizure) and Post-ictal (after seizure). The prediction task is aimed to identify preictal state from the other three states [6].

In recent years, deep learning algorithms have been used for EEG signals analysis, where the most representative algorithm is convolutional neural network (CNN). Truong et al. used Short Time Fourier Transform (STFT) and CNN with 2D convolution to process the EEG signals [7]. Eberlein et al. processed raw EEG signals in time domain with a deep CNN to predict seizure onset [8]. Xu et al. used 1D convolution for seizure prediction and got AUC and sensitivity of 0.985 and 98.2% respectively [9]. However, all these methods are computationally expensive and memory consuming, which makes it hard to deploy these methods on at-edge hardware systems where computation and memory resources are limited. Some researchers have made efforts to explore hardware friendly approaches. Truong et al. used integer CNN and binary weights CNN for seizure detection, where energy consumption per task is reduced by over 90% [10]. Zhao et al. designed a 1D Binary-CNN for seizure prediction which managed to reduce the parameter memory by 7.2 times and keep sensitivity at 0.94 [11]. However, Truong's work uses 4-bit integer weight and Zhao's work uses fully precise value in the first convolution layer and the first full-connected layer, which wastes both energy and computation resources.

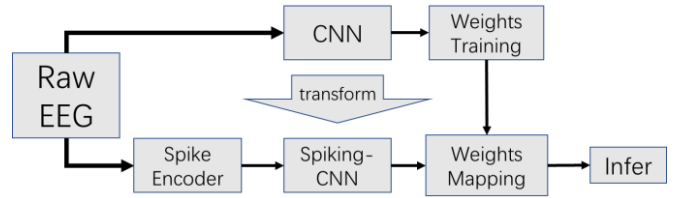

Fig. 1. Flow diagram of the proposed approach.

Biologically inspired, Spiking NN (SNN), the most representative form of neuromorphic computing, proved energy efficient and powerful [12-14]. It is expected to be the next generation of Artificial NN (ANN) [15]. Guo et al. proposed a supervised SNN with a selected spike encoder to detect interictal spikes in epilepsy based on EEG and achieved an accuracy of 92.67% [16]. Therefore, SNN can be a possible candidate for energy efficient seizure prediction. However, SNNs always use complex training methods such as unsupervised spike-timing-dependent plasticity (STDP) [17], Tempotron [18] and SpikeProp [19]. These methods are still hard to build networks as deep as CNN [20], which results in limitations to the performance and application of SNN. Spiking CNNs (Spiking-CNNs) combine the advantages of both CNN and SNN to solve above described drawbacks [20]. Weights are trained in CNN fashion and subsequently mapped to the SNN topology correspondingly [21]. A spike encoder is also used to encode the input data into spike sequences, which is specifically designed for different kinds of data and tasks. In this way, Spiking-CNN manages to keep the performance and reduce the computation complexity.

In this paper, we describe the implementation of Spiking-CNN and spike encoder used to make seizure predictions accurately and energy-efficiently. The proposed approach achieves sensitivity of 95.1% and specificity of 99.2% on raw EEG data while the computation complexity is reduced by 98.58% compared to the original CNN. The remaining sections of this paper are organized as following. Section II introduces the proposed method. Methods and materials are described in Section III. Results are shown in Section IV. Section V concludes this paper.

## II. THEORY OF THE APPROACH

The proposed approach consists of spike encoder, network models and weight mapping (Fig. 1). To predict epileptic seizures based on raw EEG, a CNN is first designed and trained to get weights, and then transformed into Spiking-CNN. To process EEG on Spiking-CNN, a spike encoder is needed to transform raw EEG to spike sequences. When weights are mapped correspondingly, the Spiking-CNN operates inferring and make the prediction.

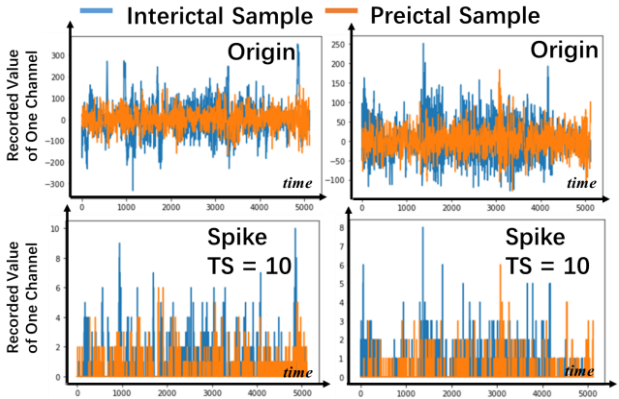

Fig. 2. Comparison between origin EEG signals and the encoded spike sequences based on 10 time-step (TS) accumulation.

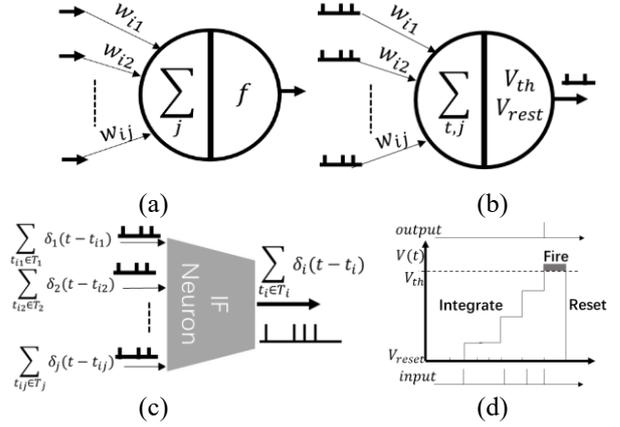

Fig. 4. Neuron Model: (a) CNN Neuron, (b) SNN IF Neuron, (c) inputs and outputs of an IF Neuron, (d) Procedure of integrating, fire and reset of IF neurons.

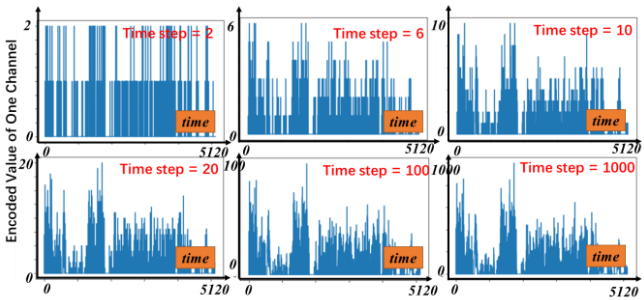

Fig. 3. Comparison among the encoded spike sequences based on different time-step accumulation.

*A. Spike-Encoder*

To convert the continuous EEG signal into discrete time-dependent spike sequences, a temporal Gaussian random sparse encoding algorithm is proposed. The encoder converts the 2D input vector into a 3D time-dependent vector. The sample of each time step is generated by the proposed Gaussian random discretization algorithm. After passing through the encoder, the features of the input vector are dispersed into each time step. For the proposed spike encoder, the *time-step*, $V_{th\text{-}up}$ and $V_{th\text{-}down}$ are determined parameters and the original data is the input EEG sample. To generate spike sequence in time domain, a random matrix is generated in every step of time, which is of the same size as the input sample and consists of gaussian random values (mean value is set to be the mean value of $V_{th\text{-}up}$ and $V_{th\text{-}down}$). Every element of the original data is then compared with the generated random value. If the random one is greater, the corresponding spike value will be set *0*; otherwise, the spike value will be set *1*. Given an input vector in the shape of [Channel, Height, Width], the output shall be [Time Step, Channel, Height, Width] consisting of *0/1*. The key benefits of encoding are: (a) to sparse the features of raw EEG in time domain so that it can be processed by Spiking-CNN and (b) to help reduce the length of processed values from 32-bit to 1-bit but still keep the main features of samples so that the goal of energy-efficiency can be achieved. changes, but later returns to $V_{rest}$ after firing the spike.

Fig. 2 shows the comparison between the origin and spike of interictal and preictal samples given the time step of 10. As shown, the spike sequences succeed to keep most of the features of the samples. Taking an interictal sample for example, Fig. 3 shows that the larger the time step is, the more data features the encoded sample retains, which also consumes more computing resources.

*B. Neuron Model Selection*

For Spiking-CNN, biology-inspired neuron models are proposed [29]. The spiking neurons can receive multiple inputs and be activated multiple times in the format of spike sequences, while the common deep learning neuron is only used once in one single computation. The function of time $V(t)$ and $V_{rest}$ are used to express the membrane and the resting potentials respectively. After the spike arrives, $V(t)$ changes, but later returns to $V_{rest}$ after firing the spike.

To propose a hardware friendly method, integrate and fire (IF) neuron model [29] is used in this study. The CNN neuron is shown in Fig. 4(a) while the IF neuron is shown in Fig. 4(b). For SNN IF neurons, the membrane potential is expressed with $V_i(t)$ and is updated at each time step as expressed in eq. (1)

$$V_i(t) = V_i(t-1) + Input(t) - Leakage \quad (1)$$

For IF neurons, $\sum \delta_j(t-t_j)$ describes the accumulation of input spike sequences (Fig. 4(c)) and the leakages are set to be $\Delta V$ as show in eq. (2)

$$V_i(t) = V_i(t-1) + \sum w_{ij} \sum \delta_j(t-t_j) - \Delta V \quad (2)$$

As shown in Fig. 4(d), when $V_i(t)$ reaches the threshold $V_{th}$, a spike will be fired at the output. Then the membrane potential $V_i(t)$ is reset to the resting potential $V_{rest}$.

*C. Network Design*

The proposed CNN model is first designed, then transferred to Spiking-CNN. For the EEG signal samples, they have thousands of elements in the time axis but only about twenty in the channel axis. Each channel records different regions of the brain, so mixing up the channel and dimension may decrease the prediction performance. Therefore, the proposed network adopts single-dimension kernel for convolution and max-pooling operations. The single-dimension convolutional kernels extract features while the max-pooling kernels keep the most significant information.

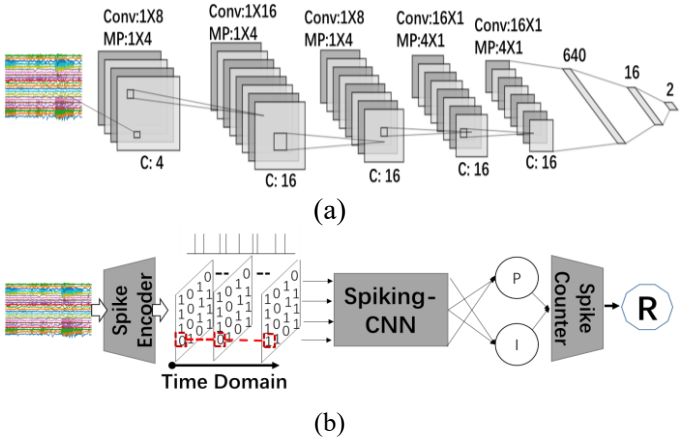

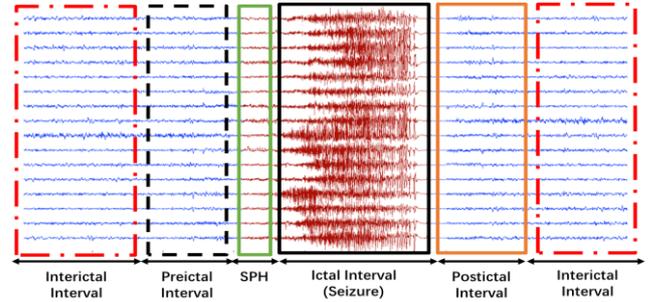

Fig. 5. Proposed Neural Network: (a) CNN structure ('Conv' is convolution, 'MP' is max-pooling, and 'C' stands for channel) and (b) Spiking-CNN structure; R, P and I stand for 'Result', 'Preictal' and 'Interictal' respectively.

Table 1. Comparison between CNN and Spiking-CNN.

| | | CNN | Spiking-CNN |
|---|---|---|---|
| Network Structure | Number of Conv | n | n |
| | Number of Max-pool | n | n |
| | Extra Layer | - | Spike Encoder, Counter |
| | Input | 2-D Vector | Spike Sequence |
| | Output | Float Value | Number of Spikes |
| Neuron Model | Weight | $w_{ij}$ | $w_{ij}$ |
| | $V_{th}$ | - | √ |
| | $V_{rest}$ | - | √ |
| | Bias | √ | √ |
| | Time Domain | - | √ |
| Operation | Convolution | MAC | ACC |
| | Max-pooling | | |
| | Activation | ReLU Function | Neuron Activation |
| Dataflow | | 32-bit Float Value | Single-bit 0 or 1 |
| Weights | | 32-bit Float Value | |

The structure of the CNN is shown in Fig. 5(a), which is adapted to achieve good performance. The proposed CNN consists of 5 convolution & max-pooling layers (Conv & MP) and 2 full-connected layers (FC Layer) plus an input and an output layer. There is a rectified linear unit (RELU) activation layer connected between each convolution layer and max-pooling layer. The corresponding structure of the Spiking-CNN is shown in Figs. 5(b). Except for the encoding layer and the spike counter (Count Layer), the transferred Spiking-CNN has the same structure as the CNN which demonstrates that Spiking-CNN uses the same feature extracting method as CNN. Spiking-CNN does not need RELU because the activation of IF neuron behaves the same as RELU [30]. The optimization lies in that all the data transferred between layers is single bit value (*0* or *1*), which transfers the former multiplication operations in CNN into adding operations. Therefore, much less computation resource is required for Spiking-CNN than CNN which makes it hardware friendly. Furthermore, the Spiking-CNN deals with data in the time domain which can be processed either sequentially or parallelly depending on the architecture

Fig. 6. Typical EEG signals in time domain.

of the computing unit. Comparison between CNN and Spiking-CNN is shown in Table 1.

*D. Procedure & Weight Mapping*

The proposed Spiking-CNN has the same structure consisting of Conv-Layer and Max-Pooling as CNN and process data in time domain (Fig. 5).

To process raw EEG data, the input samples need to be encoded into spike sequences as mentioned in Section II. Set based on the values of the recorded EEG data, $V_{th-up}$ and $V_{th-down}$ are hyperparameters of the spike encoder and are adjusted along with the chosen time step to achieve the best performance.

At every time step, the corresponding spike data is processed in the network. Every time this procedure is completed, either or both or neither of P and I neuron is activated. The spike counter in Fig. 5(b) records the number of times of activation of the two output neurons through all the time steps and greater one tells the result of prediction.

As for the weight mapping, we use SINABS [22] in this work. Due to the same structure of two networks, the trained CNN weights (32-bit floating numbers) of every layer can be restored and mapped on Spiking-CNN directly. To make it more convenient to implement the network on hardware, the resting potential of all the IF neurons are set zero. The threshold of each layer's IF neurons is adjusted along with the $V_{th-up}$ and $V_{th-down}$ of the spike encoder to improve the performance.

## III. DATA PROCESSING

*A. EEG Dataset*

To evaluate the performance of proposed neural network, we used the CHB-MIT EEG dataset, which contains scalp EEG data of 23 measurements from 22 patients [23]. All measurements are recorded at 256 Hz sampling rate using different signal acquisition settings of electrodes among patients. 15 measurements are recorded in the same fixed 23-electrode implementation, while there are some changes in electrode configuration for the remaining measurements [24].

Fig. 6 shows the different EEG zones in time domain. When dealing with seizure prediction tasks, a short time between the end of Preictal Interval and Seizure (Ictal Interval) is defined as Seizure Prediction Horizon (SPH). Preictal Interval Length (PIL) and SPH are two empirical parameters which determines the chosen intervals for the experiments. 30 min PIL and 5min SPH are chosen in this work. Then two categorized samples are extracted from Interictal Intervals and Preictal Intervals as shown in Fig. 6 with a fixed 20-second time window. For one input sample, the width is *20s*sample-rate* standing for the selected data and the height is the number of recording channels, which is *23*5120* for CHB-MIT dataset. However, the raw EEG data contains much more Interictal Intervals than Preictal Intervals causing imbalanced training sample problem which may lead the trained model to perform bad [25], [26]. To overcome this limit, preictal samples are extracted with 5-second overlapping but interictal samples are done without overlapping [9].

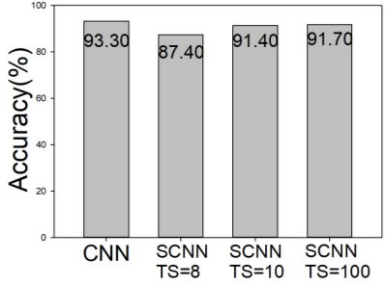

Fig. 7. Performance of accuracy. TS stands for time step.

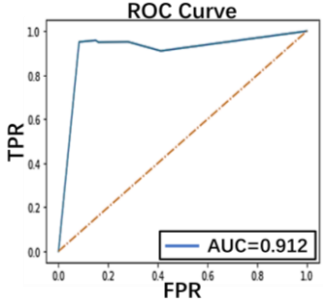

Fig. 8. ROC curve and AUC.

Table 2. ACC, TPR and FPR of different count thresholds.

| Count Threshold | ACC | TPR/Sen | FPR |
|---|---|---|---|
| 0 | 0.9141 | 0.951 | 0.0827 |
| 1 | 0.8167 | 0.9493 | 0.1608 |
| 2 | 0.8299 | 0.9582 | 0.1499 |
| 3 | 0.6247 | 0.9516 | 0.2793 |
| 4 | 0.3211 | 0.9101 | 0.4121 |

*B. Training and Inferring*

The EEG dataset is separated into 2 parts randomly with the ratio of 4:1--training set and testing set. The training set is used to train the proposed CNN. The trained weights of every layer which will be mapped on the corresponding Spiking-CNN. The testing set is used to evaluate the performance of the proposed Spiking-CNN. When making the prediction, the EEG data first goes through the spike encoder and transforms to spike sequences. Then the Spiking-CNN mapped with trained weights receives the spike sequences and give the output of prediction. By comparing the output and the label, the performance of the proposed method can be evaluated.

IV. RESULTS

In this experiment, only lead seizures occurring at least 4h after previous seizures are considered [27], so there are 7 subjects in CHB-MIT dataset suitable for experiments. In this work, the performance is the mean value of every metric of the 7 subjects. The comparison of the maximum accuracy among the proposed original CNN and Spiking-CNN with different time steps is shown in Fig. 7. The time step of 10 is sufficient for the encoder to extract features from the original data for it has achieved an accuracy of over 90%. The computation complexity of CNN is shown in eq. (3) and the computation complexity of the proposed Spiking-CNN is object of eq. (4).

Table 3. Comparison with other works on CHB-MIT dataset.

| Method | FPR(/h) | SEN(%) | AUC | Number of ADD | Number of MUL | Mem | Num of Weight |
|---|---|---|---|---|---|---|---|
| STFT-CNN [7] | 0.20 | 69.83 | _ | 1.17M | 1.45M | 3.81M | 119K |
| Wavelet-CNN [8] | _ | _ | 0.917 | 1.74M | 2.20M | 1.70M | 53.1K |
| CSP-CNN [28] | 0.12 | 92.0 | 0.90 | 1.43M | 1.97M | 2.27M | 70.9K |
| BSDCNN [11] | 0.095 | 94.69 | 0.97 | 5.54M | 0.61M | 0.067M | 67K |
| Spiking-CNN (This work) | 0.08 | 95.1 | 0.914 | 2.389M | 0 | 0.33M | 10.3K |

$$T_{CNN} = \sum M_H M_W (K_H K_W + K_H + K_W - 1) C_{in} C_{out} (1.1) \text{float-bits} \quad (3)$$

$$T_{SCNN} = \sum M_H M_W (K_H + K_W - 1) C_{in} C_{out} / 10 \text{ time-step} \quad (4)$$

where *T, M, K, C, H, W* stand for *Time Complexity, Feature Map, Kernel, Channel, Height* and *Width*. With time step chosen to be 10, the time complexity is reduced by 98.58% (including the reduction caused by the value transforming from float value to single bit value) compared with the original CNN with just a 2% loss of accuracy (Fig. 7).

Accuracy (ACC), sensitivity (TPR/Sen), false prediction rate (FPR), receiver operating characteristic (ROC) and area under curve (AUC) are also evaluated in this study. With different count thresholds, the metrics are shown in Table 2 and Fig. 8. As shown, the mean sensitivity, FPR and AUC reach 95.1%, 0.0827 and 91.2% respectively. Table 3 shows comparison of this work with other works on metrics of FPR, sensitivity, AUC as well as the estimated number of ADD operation, number of MUL operation, memory and number of weights. The comparison of overall performance with other works is based on a figure-of-merit (FOM) whose definition is given in eq. (5)

$$FOM = (SEN + AUC - FPR) / [2(10MUL + ADD + Mem)] \quad (5)$$

V. CONCLUSION

In this study, we propose a neuromorphic approach of Spiking-CNN and all the relevant modules for energy-efficient epileptic seizure prediction. By designing a temporal Gaussian random sparse encoding algorithm, we propose a spike encoder to transform raw EEG to spike sequences which can be processed in Spiking-CNN. To keep high accuracy and reduce computation resource, we design and train the CNN first and then map the weights on the Spiking-CNN using SINABS [22]. As demonstrated by the results, the proposed approach not only achieves the sensitivity and AUC of 95.1% and 0.912 respectively, but also manages to reduce computation complexity by 98.58% with just a 2% loss of accuracy.

ACKNOWLEDGMENT

The authors would like to acknowledge the financial support and tools received from Westlake University and Zhejiang Key R&D Program No. 2021C03002 to support this project.